\documentclass[wcp]{jmlr}

\usepackage{booktabs}
\usepackage{amsfonts,bbm,bm}

\DeclareMathOperator{\var}{var}
\DeclareMathOperator{\covar}{covar}
\newcommand{\Rho}{\mathrm{P}}

\jmlrworkshop{ACML2017}

\title[Quantum-Inspired Forest]{A Quantum-Inspired Ensemble Method and \\ Quantum-Inspired Forest Regressors}

  \author{\Name{Zeke Xie} \Email{xie@ms.k.u-tokyo.ac.jp} \\
  \addr The University of Tokyo, 7 Chome-3-1 Hongo, Bunkyo, Tokyo, Japan
  \AND
  \Name{Issei Sato} \Email{sato@k.u-tokyo.ac.jp}\\
  \addr The University of Tokyo/RIEKN, 7 Chome-3-1 Hongo, Bunkyo, Tokyo, Japan
 }

\editors{Y.-K. Noh and M.-L. Zhang}

\begin{document}

\maketitle

\begin{abstract}
We propose a Quantum-Inspired Subspace(QIS) Ensemble Method for generating feature ensembles based on feature selections. We assign each principal component a Fraction Transition Probability as its probability weight based on Principal Component Analysis and quantum interpretations. In order to generate the feature subset for each base regressor, we select a feature subset from principal components based on Fraction Transition Probabilities. The idea originating from quantum mechanics can encourage ensemble diversity and the accuracy simultaneously. We incorporate Quantum-Inspired Subspace Method into Random Forest and propose Quantum-Inspired Forest. We theoretically prove that the quantum interpretation corresponds to the first order approximation of ensemble regression. We also evaluate the empirical performance of Quantum-Inspired Forest and Random Forest in multiple hyperparameter settings. Quantum-Inspired Forest proves the significant robustness of the default hyperparameters on most data sets. The contribution of this work is two-fold, a novel ensemble regression algorithm inspired by quantum mechanics and the theoretical connection between quantum interpretations and machine learning algorithms.
\end{abstract}

\hyphenpenalty=5000
\tolerance=1000


\section{Background}

The goal of ensemble learning is to combine the predictions of multiple base learners to get more accurate aggregate predictions. Ensemble learning algorithms frequently rank top in many data mining competitions, and consistently outperform single learners, such as Support Vector Machines. This approach has proven to be a powerful method in practical applications, especially for those general-purpose tasks. The ensemble method generally is favored in terms of increasing robustness and accuracy. Since the theoretical analysis of ensemble models, particularly tree ensembles, has been carefully studied, we are able to theoretically analyze novel ensemble algorithms besides the empirical analysis. Many researchers have contributed to a significant amount of good works in last decades. We can find recent enhancements of Random Forest \citep{breiman2001random} in \citep{fawagreh2014random}, including perfect random tree ensembles \citep{cutler2001pert}, extremely random trees \citep{geurts2006extremely}, and completely random decision trees \citep{liu2005maximizing,fan2006general}.

Researchers have known that ensemble diversity and the accuracy of base learners are two main factors deciding the performance of ensemble models \citep{zhou2012ensemble}. And improving strength of individual trees and decreasing the correlation between trees are main factors in reducing the Random Forest error rate. We usually inject randomness into ensemble models aiming at generating diversified base learners and ensemble strategies. Unfortunately, the randomness approach generally reduced the accuracy of base learners. It's not surprising that randomness may lead some slight deviation from optimal base learners. Researchers find it quite difficult to improve ensemble diversity without damaging the accuracy of base learners. How to deal with the trade-off between diversity and accuracy becomes one of core challenges in ensemble learning.

Principal Component Analysis (PCA) \citep{abdi2010principal} is a widely used dimension reduction technique. And PCA as preprocessing is not a new thing for high dimensional regression. A classical technique named Principal Component Regression (PCR) omits the PC¡¯s corresponding to small eigenvalues and then trains regression models based on principal components. The threshold will be pretty obvious and then preserve principal components deterministically. The proposed method also tends to select components with larger eigenvalues as well as PCA. But we inject randomness into the process based on the probabilistic weights inspired by quantum mechanics.

Quantum-Inspired Machine Learning means machine learning algorithms that involve in some quantum theoretical elements but don't require a quantum machine for implementing it. Quantum physics and machine learning can be deeply interconnected in theoretical analysis. Several works of algorithms utilizing Quantum physics can be seen in literals. Quantum annealing is a most studied quantum inspired algorithm for solving combinatorial optimization problems and was proposed in \citep{kadowaki1998quantum}. Quantum annealing is inspired by the Quantum Tunneling effect to escape local optima. In recent years, multiple quantum-inspired machine learning algorithms have been proposed. For example, \citep{leifer2008quantum} reported quantum belief propagation; \citep{weinstein2009dynamic} reported a quantum-inspired clustering method; \citep{huang2012new} reported a quantum-inspired anomaly detection algorithm; \citep{blacoe2013quantum} reported a quantum-inspired semantic space model.

\subsection{Overview}
We interpret the ensemble learning process in several quantum physics concepts, and merge quantum-inspired techniques into the ensemble method naturally. We mainly focus on Tree Ensemble methods due to two truths. First, Tree Ensemble is a powerful and robust method that is widely used in multiple domain's tasks. An significant improvement on this popular method could make QIS very valuable. Second, base learners are constructed independently and in parallel. This indicates that we may take many elements of Tree Ensemble as black boxes except for generating the feature subsets. We make QI Forest and Random Forest only differ in generating feature subsets for individual learners. It provides the advantage that we can ensure any performance differences are purely caused by the proposed Quantum-Inspired Subspace method.

In Section~\ref{section2}, we present quantum interpretations (heuristics) and the proposed algorithms. We show the process how quantum mechanics inspires us to invent a novel ensemble method. In Section~\ref{section3}, we provide a solid mathematical proof for the advantage of the proposed algorithm. We prove that Quantum-Inspired Forest Regressors' advantage over Random Forest in case of the first order approximation. In our mathematical analysis, the Linear Regressor is nonlinear base regressors' first order approximation. In Section~\ref{section4}, we empirically compare Quantum-Inspired Forest Regressors and Random Forest Regressors on date sets from UCI Repository \citep{Lichman:2013}. What's more, we perform one more empirical comparison, where we take Linear Regressors as base learners instead of Decision Tree Regressor. In Section~\ref{section5}, we discuss and summary our main work.

\section{The Quantum-Inspired Approach}
\label{section2}

\subsection{Quantum Interpretations}
As the theoretical proof given in Section~\ref{section3} is sufficient, readers without quantum background may choose to skip quantum interpretations or heuristics. But quantum interpretations are especially helpful to explain how we design the proposed method at the very beginning. And the quantum interpretations provide new insights and show one step ahead of how to marry the two disciplines. Quantum physics shares similar forms with machine learning. And these similar forms or equations encourage us to think about machine learning in a quantum theoretical way. The motivation behind our works starts from Principal Component Analysis (PCA) and density matrices. \citep{nielsen2010quantum} introduced density matrix and operators in detail. In quantum mechanics, physicists often denote a pure state as a state vector $|\psi\rangle$. However, there exist mixed states, which cannot be written as a state vector. A mixed state corresponds to a probabilistic mixture of pure states, also called a quantum ensemble. A density matrix is a matrix that describes a quantum mixed state, an ensemble of several pure states. We show how to establish connections between density matrix and quantum operators to PCA as follows.

We interpret principal components as eigenstates in a mixed state. Suppose we are given a data set $X \in \mathbb{R}^{n \times m}, \vec{y} \in \mathbb{R}^{n}$ for a regression or classification problem. $X$, a $n \times m$ data matrix, contains $n$ data samples, and each feature vector $\vec{x^{i}}$ has $m$ features. The target variable vector $\vec{y}$ is a vector with a length of $n$. We define the Gram matrix $\Rho = XX^{\top}$ that is a symmetric and positive semi-definite $n \times n$ matrix. And then we have $\Rho = XX^{\top}  = U \Sigma^{P} U^{\top}$ where $\Sigma^{P}$ is a $n \times n$ diagonal matrix. Column vectors of $US$ are equal to principal components in PCA. And people often use first $k$ column features $US$ as  dimension-reduced $k$-dimension feature vectors.

The quantum journey begins from here. As the density matrix of quantum mechanics is Hermitain, positive semi-definite and of trace 1, if we normalize the Gram matrix $\Rho$ by multiplying a factor $\frac{1}{Tr(\Rho)}$, the Gram matrix can be regarded as a density matrix in quantum theory. For simplicity of our notation, we denote the normalized Gram matrix by $\rho$. And we redefine $\rho$ with a normalization factor as
\begin{align}
\rho = \frac{XX^{\top}}{Tr(XX^{\top})} = U \Sigma U^{\top}.
\end{align}
Let $\vec{u_{i}}$ denote the $i$th column vector of matrix $U$, so $\vec{u_{i}}$ is also a pure state vector, which denotes $|u_{i}\rangle$ in quantum theory. As we have replaced the Gram Matrix by the normalized $\rho$, the sum of diagonal elements of $\Sigma$, $\sum_{i=1}^{n} s_{i}^{2}$, is equal to 1. The density matrix $\rho$ describing the data matrix as a mixed state is also an operator of the form
\begin{align}
\rho = \sum_{i=1}^{n} s_{i}^{2} |u_{i} \rangle \langle u_{i} | = \sum_{i=1}^{r} s_{i}^{2} |u_{i} \rangle \langle u_{i} |
\end{align}
Physically, it means a data matrix $X$ can be regarded as a mixed state or a quantum ensemble consisting of $r$ pure states, where $r$ is the rank. In physics, an ensemble of pure states $\rho$ can reflect statistical expectations of quantum systems $|u_{i} \rangle$. And the variance $s_{i}^{2}$ is the fraction(weight probability) of the ensemble in each pure state $|u_{i} \rangle$.

On the one hand, the quantum interpretation treats PCA naturally as a dimensionality reduction process. In machine learning, researchers usually preserve the first $k$ components with largest variance values as dimensionality reduced features. In quantum mechanics, PCA means that we remove several non-principal eigenstates from the mixed state and preserve those principal eigenstates so that we prepare a new mixed state consisting of less eigenstates. The new state is exactly a low-rank approximated copy of the original mixed state. Obviously, PCA makes clear sense to us from a viewpoint of physics. But PCA is also a naive and biased operation that assigns uniform weights to principal eigenstates and weight $0$ to non-principal eigenstates.

And the second quantum interpretation is we can also regard regression as a state preparation process that we operate several pure states
$|x_{1}\rangle, |x_{2}\rangle, \dots, |x_{n}\rangle$ to approximate a target state $|y \rangle$. Translated in quantum theoretical language, it can be written as
\begin{align}
\rho_{y} = |y \rangle \langle y| = \hat{A}\rho_{x}\hat{A}^{\dag},
\end{align}
where the state operation is noted by some quantum operator $\hat{A}$. So the quantum mechanism of regression tasks can be understood as we learn a Model Operator to operate eigenstates in a mixed to approximate a target pure state under some metrics. From a quantum theoretical viewpoint, the importance of an eigenstate $|u_{i} \rangle$ is also reflected by the Transition Probability from en eigenstate $|u_{i} \rangle$ jumping into the target state $|y \rangle$. We denote Transition Probability Amplitude as $t$, so $t_{i} = \langle y|\hat{A}|u_{i} \rangle$, just like an electron jumps from one state to anther state. And Transition Probability equals Transition Probability Amplitude squared, namely $|\langle y|\hat{A}|u_{i} \rangle|^{2}$. Obviously, the Transition Probability is a parameter decided by model operator, the eigenstate, and the target state together. Aggregating fraction probabilities and transition probabilities together, the Fraction Transition Probability for the $i$th principal component is proportional to $s_{i}^{2}|\langle y|\hat{A}|u_{i} \rangle|^{2}$. So we take the the Fraction Transition Probability for the $i$th principal component as
\begin{align}
p_{k} = \frac{s_{k}^{2}t_{k}^2}{\sum_{i=1}^{r}s_{i}^{2}t_{i}^2}.
\end{align}
In Section~\ref{section3}, we prove that Transition Probabilities happen to equal parameters squared of linear regression mapping from $X$ to $y$ in the first order approximated situation. According to the heuristical Fraction Transition Probabilities, we successfully propose Quantum-Inspired Subspace Method and Quantum-Inspired Forest.

\subsection{Algorithm}

\begin{algorithm2e}
\DontPrintSemicolon
\SetKwInOut{Input}{Input}
\SetKwInOut{Output}{Output}
\underline{function QISubspace} $(X, y, \set{F}, T, K)$\;
\Input{the data matrix $X$, the target variable vector $\vec{y}$, the ensemble size $T$, the target space dimensionality $K = \alpha m$}
\Output{feature subsets $\{\set{F}_{i}| i=1, \dots,T\}$}
Preprocess data matrix $X_{R} \gets PCA(X)$ by using full-rank PCA \;
Compute Fraction Probabilities $\vec{p_{s}} \gets$ the diagonal elements of covariance matrix $X^{\top}X$\;
Compute Transition Probability Amplitudes $\vec{t} \gets (X_{R}^{\top}X_{R})^{-1}X_{R}^{\top}\vec{y}$ which are LR parameters\;
Compute Transition Probabilities $\vec{p_{t}} \gets \vec{t}.*\vec{t}$\;
Compute Fraction Transition Probabilities $\vec{p} \gets \frac{\vec{p_{s}}.*\vec{p_{t}}}{norm(\vec{p_{s}}.*\vec{p_{t}})}$\;
\For{$i \gets 1$ \textbf{to} $T$} {
  Select K unique random integers $a_{1}, \dots, a_{K}$ from $[1, m]$ in probabilities of $p_{a_{i}}$\;
  $\set{F}_{i} \gets \{a_{1}, \dots, a_{K}\}$\;
}
\Return{$\{\set{F}_{i}| i=1, \dots,T\}$}
\caption{Quantum-Inspired Subspace for generating feature subsets}
\label{algo:QIS}
\end{algorithm2e}

\begin{algorithm2e}
\DontPrintSemicolon
\SetKwInOut{Input}{Input}
\SetKwInOut{Output}{Output}
\underline{function RandomForest} $(S, \set{F}, T, K)$\;
\Input{A training set $S = (x^{1}, y^{1}), . . . ,(x^{n}, y^{n})$, features $\set{F}$, and the forest size $T$, the target space dimensionality $K$}
\Output{Random Forest $H$}
 $H \gets \emptyset$\;
 \For{$i \gets 1$ \textbf{to} $T$} {
  $S^{i} \gets$ a bootstrap sample from $S$\;
  $\set{F}^{i} \gets$ a random subset of size $K$ sampled from $\set{F}$\;
  $h_{i} \gets$ TreeLearn($S^{i}, \set{F}^{i}$)\;
  $H \gets H \cup \{h_{i}\}$\;
 }
 \Return{H}\;
\caption{Random Forest}
\label{algo:RF}
\end{algorithm2e}

\begin{algorithm2e}
\DontPrintSemicolon
\SetKwInOut{Input}{Input}
\SetKwInOut{Output}{Output}
\underline{function QIForest} $(S, \set{F}, T, K)$\;
\Input{A training set $S = (x^{1}, y^{1}), . . . ,(x^{n}, y^{n})$, features $\set{F}$, and the forest size $T$, the target space dimensionality $K$}
\Output{Quantum-Forest $H$}
 $H \gets \emptyset$\;
 $\{\set{F}_{i}| i=1, \dots,T\}$ generated by \underline{function QISubspace} $(X, y, \set{F}, T, K)$\;
 \For{$i \gets 1$ \textbf{to} $T$} {
  $S^{i} \gets$ a bootstrap sample from $S$\;
  $\set{F}^{i} \gets \set{F}_{i}$\;
  $h_{i} \gets$ TreeLearn($S^{i}, \set{F}^{i}$)\;
  $H \gets H \cup \{h_{i}\}$\;
 }
 \Return{H}\;
\caption{Quantum-Inspired Forest}
\label{algo:QIF}
\end{algorithm2e}

Random Subspace is a fast and efficient ensemble method widely used in many algorithms, including Random Forest. Random Subspace randomly select a subset of features for training a base learner. But Quantum-Inspired Subspace can utilize the extra information inspired by quantum mechanics. We first preprocess the input data matrix $X$ by using full-rank PCA. Different from either preserving principal components with largest eigenvalues or random subspace, QIS selects a component in a probability proportional to the corresponding Fraction Transition Probability. Under Gaussian assumptions of model parameters, we let $p_{k} = \frac{s_{k}^{2}}{\sum_{i=1}^{r}s_{i}^{2}}$ for the component $k$. When we replace Random Subspace by Quantum-Inspired Subspace for Random Forest, we obtain a novel algorithm, namely Quantum-Inspired Forest. We note that, in principle, full-rank PCA preprocessing generally can neither improve nor damage algorithm performance. The additional computational cost of the proposed algorithm is only brought by Principal Component Analysis and several matrix operations for computing Fraction Transition Probabilities. So it is a very low cost in practice.

Denote by $h_{1}, \dots, h_{T}$ the regressors in the ensemble and by $\set{F}$, the feature set. As with most ensemble methods, we need to choose ensemble size $T$ in advance. All base regressors can be trained in parallel, which is also the case with Bagging and Random Forests. Algorithm~\ref{algo:QIS} explains how to generate construct the training feature set $\set{F}_{i}$ for regressor $h_{i}$. And we modify Random Forest into Quantum-Inspired Forest by employing Quantum-Inspired Subspace to generate ensemble feature subsets instead of Random Subspace. We can easily notice the difference between standard Random Forest and Quantum-Inspired Forest respectively described in Algorithm~\ref{algo:RF} and Algorithm~\ref{algo:QIF}.

It is worthy noting that Quantum-Inspired Subspace is a general method which can be easily applied with other ensemble methods and multiple base learners together. QIS also lend itself naturally to parallel processing, as ensemble feature sets and individual learners can be built in parallel. QIS is not only naturally applicable to Tree Ensembles, but also makes sense for any ensemble regressors whose diversity is based random feature selections.

\section{Theoretical Analysis and Proof}
\label{section3}

In this section, we prove the advantage of Quantum-Inspired Subspace through error-variance-covariance decomposition that combines error-ambiguity decomposition and bias-variance-covariance decomposition together. The proof states that the advantage of QIS theoretically increase ensemble ambiguity and decrease the individual error expectation in the first order approximation. And in our empirical analysis, the experimental results support the advantage is still approximately applicable to nonlinear models, such as Decision Tree. The mathematical proof for ensemble classification cannot hold in the same way, although our empirical analysis support that Quantum-Inspired Forest Classifiers can be favorably compared with Random Forest Classifiers.

\subsection{Error-Variance-Covariance Decomposition}
In this section, we show how to obtain Error-Variance-Covariance Decomposition. We organize several known conclusions together referring to derivations in Chapter 5.2 of \citep{zhou2012ensemble}.
Assume that the task is to use an ensemble of T base regressors $h_{1}, h_{2}, ..., h_{T}$ to approximate a function $f: \mathbb{R}^{m} \rightarrow \mathbb{R}$. And a simple averaging policy is used for the final ensemble prediction
\begin{align}
H(\vec{x}) = \frac{1}{T} \sum_{i=1}^{T}h_{i}(\vec{x}),
\end{align}
where $H(\vec{x})$ is the ensemble learner. And we define several notations here. The generalization error and ambiguity of a base learner is respectively defined as
\begin{align}
err(h_{i}) = (h_{i}(x)-f(x))^2 , \\
ambi(h_{i}) = (h_{i}(x)-H(x))^2.
\end{align}
And we also note the expectation prediction of a base learner $h_{i}$ as
\begin{align}
\mathbb{E}[h_{i}] = \int h_{i}(x) p(x) dx,
\end{align}
where $p(x)$ is the density function for data $x$.
On the one hand, \citep{krogh1995neural} proposed the error-ambiguity decomposition of ensemble learning, and the generalization error of the ensemble can be written as
\begin{align}
err(H) = \overline{err}(H) -\overline{ambi}(H) ,
\label{eq:errambi}
\end{align}
where $\overline{err}(H)= \frac{1}{T} \sum_{i=1}^{T} err(h_{i})$ is the average of individual generalization errors, and $\overline{ambi}(H)= \frac{1}{T} \sum_{i=1}^{T} ambi(h_{i})$ is the average of ambiguities which is also called the ensemble ambiguity. A basic truth is that the larger the ensemble ambiguity, the better the ensemble.

On the other hand, \citep{ueda1996generalization} developed the bias-variance-covariance decomposition. The averaged bias, averaged variance, and averaged covariance of the individual learners are defined respectively as
\begin{align}
\overline{bias}(H)  =   \frac{1}{T} \sum_{i=1}^{T} (\mathbb{E}[h_{i}]-f) ,
\end{align}
\begin{align}
\overline{variance}(H)  =  \frac{1}{T} \sum_{i=1}^{T} \mathbb{E}[(h_{i}-\mathbb{E}[h_{i}])^2] ,
\label{eq:var}
\end{align}
\begin{align}
\overline{covariance}(H)  = \frac{1}{T(T-1)} \sum_{i=1}^{T} \sum_{j \neq i, j=1}^{T} \mathbb{E}[(h_{i} - \mathbb{E}[h_{i}])(h_{j}-\mathbb{E}[h_{j}])].
\label{eq:covar}
\end{align}
And then the bias-variance-variance decomposition of ensemble is written as
\begin{align}
err(H) = \overline{bias}(H)^{2} + \frac{1}{T}\overline{variance}(H)+(1-\frac{1}{T})\overline{covariance}(H).
\end{align}
We may establish a bridge connecting the error-ambiguity decomposition and the bias-variance-covariance decomposition\citep{brown2005managing,brown2005diversity} as
\begin{align}
\overline{err}(H) - \overline{ambi}(H) = \overline{bias}(H)^{2} + \frac{1}{T}\overline{variance}(H)+(1-\frac{1}{T})\overline{covariance}(H).
\end{align}
And then we have
\begin{align}
\overline{err}(H) &= \mathbb{E}\left[\frac{1}{T} \sum_{i=1}^{T} (h_{i}-f)^{2}\right] = \overline{bias}^{2}(H) + \overline{variance}(H),
\end{align}
\begin{align}
\overline{ambi}(H)  &= \mathbb{E}\left[\frac{1}{T} \sum_{i=1}^{T} (h_{i}-H)^{2}\right] \nonumber \\
                    &= (1 - \frac{1}{T})\overline{variance}(H) - (1-\frac{1}{T})\overline{covariance}(H).
\end{align}
Finally, we obtain Error-Variance-Covariance Decomposition as
\begin{align}
err(H)  = \overline{err}(H) - (1-\frac{1}{T})\overline{variance}(H) + (1-\frac{1}{T}) \overline{covariance}(H).
\end{align}
And the generalization error expectation is written as
\begin{align}
\mathbb{E}[err(H)]  = \mathbb{E}[err(h_{i})] - (1-\frac{1}{T})\mathbb{E}[\var(h_{i})] + (1-\frac{1}{T}) \mathbb{E}[\covar(h_{i}, h_{j})].
\end{align}
Actually, there is no simple ensemble method that can minimize the expectation of $err(H)$. Fortunately, according to our following analysis, we find Quantum-Inspired Subspace method can decrease $\mathbb{E}[err(h_{i})]$, $ - \mathbb{E}[\var(h_{i})]$ and $\mathbb{E}[\covar(h_{i}, h_{j})]$ simultaneously, compared to Random Subspace method.

\subsection{Ensemble Ambiguity}

We decide to prove that Quantum-Inspired Subspace can improve ensemble ambiguity $\overline{ambi}(H)$ and decrease individual generalization errors $\overline{err}(H)$ simultaneously. And according to the Error-Variance-Covariance Decomposition relations, increasing ensemble ambiguity is equivalent to increasing $\mathbb{E}[\var(h_{i})] - \mathbb{E}[\covar(h_{i}, h_{j})]$. We want to figure out how to improve $\overline{variance}(H) - \overline{covariance}(H)$. We note that nonlinear regression models degenerate to Linear Regression (LR) in case of the first order approximation, just like how Taylor series expansion works. In the case of the first order approximation, we ignore all high order nonlinear terms. And we find the approximated case holds well for regression trees, as regressions tree also aim at finding linear relationships between features and target variables.

So in this subsection, what we decide to prove actually is, with Linear Regressors as base regressors, Quantum-Inspired Subspace Ensemble method can increase ensemble ambiguity strictly. Although it seems naive to consider ensemble linear regressors only, the mathematical analysis provides important theoretical insights about other nonlinear base learners. Assuming model parameters are independent distributed Gaussian random variables, we further know QIS can even decrease the averaged individual generalization errors. Given general data sets instead a certain data set, the Gaussian assumption that takes model parameters as Gaussian random variables is reasonable and realistic for most machine learning models. But the independence assumption only approximately holds for several linear models, luckily including Linear Regression. However, although what we prove only holds for most simplified cases, we find the proof still partly holds in more general situation. For simplicity, we use several new notations in proof. We denote the original data matrix as $X^{\prime}=U S V^{\top}$ and its linear regression parameters as $w_{k}^{\prime} \sim \bm{N}(0,\sigma^{2})$ , where $k=1, \dots, m$. We can safely assume each parameter independently obeys normal distribution as we have no prior knowledge about the importance of features. Considering a certain data set, without training, we of course know nothing about each feature's importance. Considering model performance on general data sets, the independent Gaussian assumption is also realistic.

Let's turn to the full-rank PCA preprocessed data matrix $X=X^{\prime}V$ and its linear regression parameters $\vec{w}=V^{\top}\vec{w^{\prime}}$. As $V$ is an orthogonal matrix, a model parameter $w$ still obeys a Gaussian distribution, $w \sim \bm{N}(0,\sigma^2)$. We may regard columns vectors of preprocessed matrix X as input features. So we define individual learners as
\begin{align}
h_{i}(\vec{x}) = \sum_{k \in \set{F}_{i}} w_{k}s_{k}u_{k} = \sum_{k \in \set{F}_{i}} w_{k}x_{k},
\label{eq:lr}
\end{align}
where $s_{k}$ is the $k$th-largest singular value, and $\set{F}_{i}$ is the feature subset for the $i$th base learner.. Benefitting from orthogonalized preprocessing and LR as base learners, model parameters stay invariant even trained by variant feature subsets. We call this characteristic as Parameter Invariance under variant feature subsets. In our proof, the Parameter Invariance of base learners is a key prerequisite for improving ensemble ambiguity. And besides Linearity, how Parameter Invariance is approximately applicable to nonlinear models is another key factor deciding how generally the proof may hold.

We first analyze the ensemble ambiguity $\overline{ambi}(H)$ which is equivalent to $(1-\frac{1}{T})(\overline{variance}(H)-\overline{covariance}(H))$.
According to Equation~\ref{eq:lr}, we have
\begin{align}
\mathbb{E}[\covar(h_{i}, h_{j})] = \mathbb{E}\left[\covar(\sum_{k\in \set{F}_{i}}w_{k}s_{k}u_{k}, \sum_{k\in \set{F}_{j}}w_{k}s_{k}u_{k})\right] = \sum_{k=1}^{r} w_{k}^{2}s_{k}^{2}p_{k}^{2},
\end{align}
\begin{align}
\mathbb{E}[\covar(h_{i})] = \mathbb{E}\left[\covar(\sum_{k\in \set{F}_{i}}w_{k}s_{k}u_{k}, \sum_{k\in \set{F}_{i}}w_{k}s_{k}u_{k})\right] = \sum_{k=1}^{r} w_{k}^{2}s_{k}^{2}p_{k}
\end{align}
with a constraint of $\sum_{k=1}^{r} p_{k}=1$ and a statistical assumption that $w_{k} \sim \bm{N}(0, \sigma^{2})$ is a normal random variable. We note that Random Subspace just naively sets $p_{k}=\frac{1}{r}$. We have a better solution to increase the ensemble ambiguity. We find the solution
\begin{align}
p_{k} = \frac{w_{k}^{2}s_{k}^{2}}{\sum_{i=1}^{r}w_{i}^{2}s_{i}^{2}},
\end{align}
which can exactly minimize $\mathbb{E}[\covar(h_{i},h_{j})]$. What's more, it further increases $\mathbb{E}_{QI}[\var(h_{i})]$ compared with $\mathbb{E}_{RS}[\var(h_{i})]$,
\begin{align}
\sum_{k=1}^{r} w_{k}^{2}s_{k}^{2}p_{k} > \sum_{k=1}^{r} \frac{w_{k}^{2}s_{k}^{2}}{r}.
\end{align}
So we have
\begin{align}
\mathbb{E}_{QI}[\var(h_{i})] > \mathbb{E}_{RS}[\var(h_{i})],
\end{align}
\begin{align}
\mathbb{E}_{QI}[\covar(h_{i}, h_{j})] < \mathbb{E}_{RS}[\covar(h_{i}, h_{j})].
\end{align}
The solution we find is in same forms as the Fraction Transition Probability that the density matrix interpretation indicates. The Transition Probabilities of Linear Regression Quantum Operator are exactly the linear regression model parameters. For a certain data set, we can get certain weights $\vec{w}$. For general data sets, we still have normal distribution assumption so that $\frac{w_{k}^{2}}{s^{2}} \sim \chi(1)$ is a chi-squared random variable. \citep{provost1994exact} revealed the analytical probability density function of $p_{k}$, and we know its expectation must be $\hat{p_{k}} = \frac{s_{k}^{2}}{\sum_{i=1}^{r}s_{i}^{2}}$, which are exactly Fraction Probabilities given by quantum interpretations. Our theoretical analysis of Quantum-Inspired Subspace shows that
\begin{align}
\mathbb{E}_{QI}[\overline{ambi}(H)] > \mathbb{E}_{RS}[\overline{ambi}(H)].
\label{eq:ambi}
\end{align}

\subsection{Individual Errors}
In this subsection, we want to explain that Quantum-Inspired Subspace, $p_{k} = \frac{w_{k}^{2}s_{k}^{2}}{\sum_{i=1}^{r}w_{i}^{2}s_{i}^{2}}$, tends to decrease the averaged individual error, namely $\overline{err}(H)$. Actually this conclusion is trivial. Although for a certain data set, we cannot conclude that each original feature equally contributes to the model performance. But, for general data sets, under the Gaussian assumption of model parameters $\vec{w}$, we can safely say that the expectation contribution of each original feature tends to be equal. As we have preprocessed data sets by using full-rank PCA, the widely accepted prior belief that principal components with larger variance carry more information supports the conclusion that QIS decreases the individual errors. As we know $\overline{err}(H) = \frac{1}{T} \sum_{1}^{T} err(h_{i}) =  \mathbb{E}[(h_{i}-f)^2]$, the widely accepted prior belief can be written as
\begin{align}
\mathbb{E}_{QI}[(h_{i}-f)^2]] & > \mathbb{E}_{RS}[(h_{i}-f)^2]] \nonumber \\
\mathbb{E}_{QI}[\overline{err}(H)] & > \mathbb{E}_{RS} [\overline{err}(H)].
\label{eq:baseerr}
\end{align}
According to Equation ~\ref{eq:errambi}, \ref{eq:ambi} and \ref{eq:baseerr}, we finally prove the conclusion that
\begin{align}
\mathbb{E}_{QI} [err(H)] < \mathbb{E}_{RS} [err(H)].
\end{align}
The proof indicates that the correlation between base learners is decreased with an expectation enhancement in their strength. Statistically speaking, QIS can even improve base learners' performance and ensemble ambiguity simultaneously. For the individual error expectation, the quantum-inspired weighted probabilistic selection strategy tends to work at least the same good as the uniform probabilistic selection strategy. We also note that this conclusion is statistically correct but not guaranteed on some certain data set.

Although Transition Probabilities of nonlinear models are quite difficult to derive, the Gaussian assumption is always realistic. We argue that Fraction Probabilities are at least approximately applicable to most machine learning models. We conjecture that, even if without Model Transition Probabilities, Fraction Probabilities are still very likely to improve ensemble learners, including classifiers.

Besides the simplified case of linear regression, we also need to discuss how Decision Tree may approximately preserve the first order linearity approximation and Parameter Invariance under variant feature subsets. On the one hand, the strategy to find the best split for constructing a regression tree is based on the criteria of mean square error reduction. So the feature split order can stay approximately invariant under variant feature subsets, whose mechanism is close to Parameter Invariance under variant feature subsets. On the other hand, the Decision Tree regressors learn linear relationships between features and target variables. The regression function based a tree regression mapping from $X$ to $\vec{y}$ can be very simple like a combination of $N$ step functions. In the limit of $N \rightarrow +\infty$, a combination of $N$ step functions tends to become a approximately smooth function. The first order approximation makes sense in this situation.

\section{Empirical Analysis}
\label{section4}

\begin{table*}[t]
\centering
\caption{QI Forest Regressors vs. Random Forest Regressors: $\alpha=0.5$; ensemble size $T=30$; training instances $N=60\%$. Mean square errors (with standard deviations as subscripts) are presented.}
\resizebox{\columnwidth}{!}{
\begin{tabular}{ccc|cc|cc}
\toprule
\textbf{Data} & \textbf{Instances} & \textbf{Dimension}  & \textbf{QI-Forest} & \textbf{R-Forest} & \textbf{$+/-$}  \\
\midrule
 Abalone                           & 4177   & 8   & $ 0.3204_{0.0055}$ & $0.3350_{0.0073}$  & $++$ \\
 Communities Crime                 & 1994   & 122 & $ 0.2763_{0.0025}$ & $0.3016_{0.0080}$  & $++$ \\
 Communities Crime Unnormalized 1  & 2215   & 140 & $ 0.2515_{0.0053}$ & $0.2766_{0.0112}$  & $++$ \\
 Communities Crime Unnormalized 2  & 2215   & 140 & $ 0.2125_{0.0052}$ & $0.2697_{0.0073}$  & $++$ \\
 Facebook Metrics                  & 500    & 11  & $ 0.1580_{0.0302}$ & $0.1267_{0.0480}$  & $-$ \\
 Forests Fire                      & 517    & 8   & $ 0.8296_{0.0175}$ & $0.8369_{0.0231}$  & $+$ \\
 Housing                           & 505    & 13  & $ 0.2011_{0.0089}$ & $0.2492_{0.0171}$  & $++$ \\
 Slump Test                        & 103    & 9   & $ 0.1704_{0.0103}$ & $0.2678_{0.0276}$  & $++$ \\
 Wine Quality Red                  & 1599   & 11  & $ 0.4379_{0.0060}$ & $0.4622_{0.0118}$  & $++$ \\
 Wine Quality White                & 4898   & 11  & $ 0.4056_{0.0025}$ & $0.4087_{0.0075}$  & $+$ \\
 \bottomrule
\end{tabular}
}
\label{table:forest}
\end{table*}

\begin{table*}[t]
\centering
\caption{QI Ensemble Linear Regressor vs. Random Ensemble Linear Regressors: $\alpha=0.5$; ensemble size $T=30$; training instances $N=60\%$.}
\resizebox{\columnwidth}{!}{
\begin{tabular}{ccc|cc|cc}
\toprule
\textbf{Data} & \textbf{Instances} & \textbf{Dimension}  & \textbf{QIE-LR} & \textbf{RE-LR} & \textbf{$+/-$}  \\
\midrule
 Abalone                           & 4177   & 8   & $ 0.3466_{0.0061}$ & $0.4186_{0.0207}$  & $++$ \\
 Communities Crime                 & 1994   & 122 & $ 0.2398_{0.0021}$ & $0.3220_{0.0275}$  & $++$ \\
 Communities Crime Unnormalized 1  & 2215   & 140 & $ 0.0213_{0.0001}$ & $0.1935_{0.0226}$  & $++$ \\
 Communities Crime Unnormalized 2  & 2215   & 140 & $ 0.1104_{0.0022}$ & $0.2389_{0.0202}$  & $++$ \\
 Facebook Metrics                  & 500    & 11  & $ 0.0044_{0.0004}$ & $0.0675_{0.0196}$  & $++$ \\
 Forests Fire                      & 517    & 8   & $ 0.7298_{0.0016}$ & $0.7332_{0.0029}$  & $++$ \\
 Housing                           & 505    & 13  & $ 0.2695_{0.0026}$ & $0.3883_{0.0247}$  & $++$ \\
 Slump Test                        & 103    & 9   & $ 0.1075_{0.0042}$ & $0.2624_{0.0446}$  & $++$ \\
 Wine Quality Red                  & 1599   & 11  & $ 0.4764_{0.0023}$ & $0.4833_{0.0103}$  & $++$ \\
 Wine Quality White                & 4898   & 11  & $ 0.5245_{0.0010}$ & $0.5334_{0.0073}$  & $++$ \\
\bottomrule
\end{tabular}
}
\label{table:lr}
\end{table*}

\begin{table*}[t]
\centering
\caption{QI Forest Regressors vs. Random Forest Regressors: ensemble size $T=30$; training instances $N=60\%$; adjust $\alpha$ respectively as $0.125, 0.25, 0.5, 0.75, 1.0$. When $\alpha=1.0$, QI Forest degenerates into Random Forest. MSE averaged over 10 datasets are presented.}
\begin{tabular}{c|cc}
\toprule
\textbf{$\alpha$}  & \textbf{QI-Forest} & \textbf{R-Forest}  \\
\midrule
0.125     &  $0.4251_{0.0154}$ &  $0.4932_{0.0208}$ \\
0.25      &  $0.3411_{0.0082}$ &  $0.4186_{0.0182}$ \\
0.5       &  $0.3263_{0.0094}$ &  $0.3544_{0.0168}$ \\
0.75      &  $0.3253_{0.0099}$ &  $0.3313_{0.0118}$ \\
1.0       &  $0.3377_{0.0095}$ &  $-$      \\
\bottomrule
\end{tabular}
\label{table:alpha}
\end{table*}

\begin{table*}[t]
\caption{QI Forest Regressors vs. Random Forest Regressors: $\alpha=0.5$; training instances $N=60\%$; adjust ensemble size $T$ respectively as $3, 10, 30, 100$.}
\centering
\begin{tabular}{c|cc}
\toprule
\textbf{$T$}  & \textbf{QI-Forest} & \textbf{R-Forest}  \\
\midrule
3      &  $0.4212_{0.0313}$ &  $0.4758_{0.0613}$ \\
10     &  $0.3565_{0.0219}$ &  $0.3888_{0.0317}$ \\
30     &  $0.3263_{0.0094}$ &  $0.3534_{0.0168}$ \\
100    &  $0.3140_{0.0046}$ &  $0.3356_{0.0076}$ \\
\bottomrule
\end{tabular}
\label{table:T}
\end{table*}

\begin{table*}[t]
\caption{QI Forest Regressors vs. Random Forest Regressors: $\alpha=0.5$; ensemble size $T=30$; adjust training instances $N$ respectively as $30\%, 40\%, 50\%, 60\%$.}
\centering
\begin{tabular}{c|cc}
\toprule
\textbf{Training Instances}  & \textbf{QI-Forest} & \textbf{R-Forest}  \\
\midrule
$30\%$   &  $1.4372_{0.0359}$ &  $1.3462_{0.0555}$  \\
$40\%$   &  $0.8310_{0.0208}$ &  $0.8879_{0.0334}$  \\
$50\%$   &  $0.5551_{0.0131}$ &  $0.6209_{0.0243}$  \\
$60\%$   &  $0.3263_{0.0094}$ &  $0.3534_{0.0168}$  \\
\bottomrule
\end{tabular}
\label{table:N}
\end{table*}

Quantum-Inspired Subspace is easily incorporated into existing algorithms. In order to examine the benefit of QIS to ensemble performance, we modify standard Random Forest to incorporate Quantum-Inspired Subspace before the tree induction phase. Random Forest and Quantum-Inspired Forest are implemented respectively according to Algorithm~\ref{algo:RF} and Algorithm~\ref{algo:QIF}. In our empirical study of Quantum-Inspired Forest and Random Forest, we selected 10 UCI data sets that are commonly used in the machine learning literature in order to make the results easier to interpret and compare. As we take LR as base learners in our proof, we also compare Random Ensemble LR with Quantum-Inspired Ensemble LR in Table~\ref{table:lr}, where we replace Decision Tree by Linear Regression as base learners. Ensemble Linear Regressors are not useful in practice, but it can show how our proof holds.

We take the averaged mean square error (MSE) on 10 data sets as the metrics in our empirical analysis. We decide to preprocess data sets, and take full-rank PCA preprocessed data matrix and mean normalized target variables $y$ as preprocessed data sets. The first purpose is to ensure any performance differences are purely caused by the proposed Quantum-Inspired Subspace method rather than full-rank PCA preprocessing. We must leave the difference from full-rank PCA out. The second purpose is to remove the scale differences of different data sets so that we can fairly evaluate overall performance on 10 data sets. It's reasonable to start from full-rank PCA preprocessing because full-rank PCA is only an orthogonal transformation and causes no loss or distortion of information. As we mentioned above, in principle, full-rank PCA generally can neither improve nor damage algorithm performance. In practice, full-rank PCA usually brings in uncertain performance improvement or damage. So the full-rank PCA preprocessing is necessary for removing the uncertain performance differences from the orthogonal transformation.

We present mean square errors (with standard deviations as subscripts) on each data set or the averaged MSE on all 10 data sets in following tables. In Table~\ref{table:forest} and ~\ref{table:lr}, we denote better, significantly better, worse and significantly worse respectively as $+, ++, -$ and $--$. Instances is the data sample size. Dimension is the original data space dimensionality. We typically take $60\%$ data instances as training data. As we notice the performance of Random Forest and Quantum-Inspired Forest adapt to hyperparameters in similar patterns, we decide to study two Forests' performance in multiple settings of forest hyperparameters. We employ the strategy to compare Quantum-Inspired Forest and Random Forest counterpart in the same hyperparameter settings. This strategy removes the performance differences from tuning hyperparameters. And we repeat each experiment for 15 times to get statistically reliable results. The hyperparameter setting for Decision Tree base learners is always fixed in our experiments. The function to measure the quality of a split is mean square error. And a tree always find the best split at each node. And we also set no tree depth limit, and no minimum samples limit for splits and leaves. The default hyperparameter setting for forests is: ensemble size $T=30$; select one half features to train base learners, which means $\alpha = 0.5$; the sub-sample size in Bagging is always the same as the original training sample size but the samples are drawn with replacements; and $N=60\%$ samples are used as training data instances.

Both Table~\ref{table:forest} and ~\ref{table:lr} share the defualt hyperparameter setting: $T=30, \alpha=0.5, N=60\%$. We present MSE and standard deviations on each data set. Table~\ref{table:alpha} shares the default hyperparameter setting except that we set $\alpha$ respectively to $0.125, 0.25, 0.5, 1.0$. In this experiment, we want discover how QI Forest is compared to Random Forest with variant $\alpha$ settings. Table~\ref{table:T} shares the default hyperparameter setting except that we set ensemble size $T$ respectively as $3, 10, 30, 100$. In this experiment, we want to discover how robustly QI Forest and Random Forest perform with small ensemble sizes. Table~\ref{table:N} shares the default hyperparameter setting except that we set training instances $N$ respectively to $30\%, 40\%, 50\%, 60\%$. In this experiment, we want to how robustly QI Forest and Random Forest solve small data problems.

Table~\ref{table:forest} shows the significant advantage of QI Forest Regressors in the default hyperparameter setting. QI Forest significantly outperform Random Forest on seven data sets; QI Forest slightly outperform Random Forest two data sets; and QI Forest perform slightly worse than Random Forest on only one data sets. The experimental result supports that Quantum-Inspired Forest Regressors outperform Random Forest Regressors in general situation. Table~\ref{table:lr} further supports our theoretical analysis in first order approximation. QI Ensemble Linear Regressors significantly outperform Random Ensemble Linear Regressors on all 10 data sets. Table~\ref{table:alpha} supports that QI Forest not only outperforms Random Forest in one setting of $\alpha$, but also beat Random Forest with multiple $\alpha$ settings. We notice that the smaller $\alpha$ is, the larger the advantage of QI Forest is. Especially when we select only a small number of features for training base learners, QI Forest can outperform Random Forest significantly. Table~\ref{table:T} indicates that the performance difference of QI Forest and Random Forest increases as the ensemble size $T$ decrease. It means QI Forest can perform significantly better than Random Forest with a limited forest size. Table~\ref{table:N} shows another advantage that QI Forest can solve small sample regression problems better than Random Forest. As we decrease training instances from $60\%$ to $30\%$, the performance difference increases significantly. These experimental results show that given very limited computational resources or training data, QI Forest can outperform Random Forest.

As for classification tasks, our preliminary empirical analysis shows similar and weaker advantage of QI Forest Classifiers. However, due to the lack of theoretical support for QI Forest Classifiers, we don't include the analysis about Quantum-Inspired Forest Classifiers in this paper. It remains to be further studied.

\section{Discussion and Conclusion}
\label{section5}

From a heuristical viewpoint, we propose novel quantum interpretations for machine learning. On the one hand, we interpret eigenvalues of PCA as Fraction Probabilities in a mixed state. And it naturally indicates a generally accepted belief that eigenvalues / Fraction Probabilities can reflect the importance of each principal component. And it becomes natural to let the probability of selecting a component be proportional to the corresponding Fraction Transition Probability. However, considering our theoretical proof is only the first order approximately applicable to ensemble regressors, we only claim the advantage of QI Forest Regressors in this paper.

From a viewpoint of theoretical analysis, we prove Fraction Probabilities and Transition Probabilities indeed can decrease ensemble errors in the simplified situation. According to our mathematical proof, in the case of Linear Regression as base learners, Transition Probabilities are exactly equal to model parameters of LR. For complex machine learning models, models' Transition Probabilities are quite difficult to derive. But for ensemble regressors, Transition Probabilities still make sense in the first order linearity approximation. As the Gaussian assumption of model parameters is almost always realistic, we argue that Fraction Probabilities are approximately applicable to forest regressors. We conjecture that, even without Model Transition Probabilities, Fraction Probabilities are still likely to improve ensemble learning.

From a viewpoint of empirical analysis, our experiments strongly support the advantage of Quantum-Inspired Forest Regressors in multiple hyperparameter settings. In Table~\ref{table:lr}, we take Linear regression as base regressors, Quantum-Inspired Ensemble Linear Regressors significantly outperform Random Linear Regressors on all 10 data sets. In other tables, we take Decision Tree as base regressors, Quantum-Inspired Forest Regressors still outperform Random Forest Regressors significantly in variant hyperparameter settings. And we can ensure any performance differences are purely caused by the proposed Quantum-Inspired Subspace Method. Our empirical analysis concludes that Quantum-Inspired Forest perform more robustly than Random Forest, given very limited computational resources or training data. The observation provides QI Forest an extra advantage in limited resources.

In summary, we have two fold of contributions. First, we propose a novel ensemble method named Quantum-Inspired Subspace and Quantum-Inspired Forest. Quantum-Inspired can be easily applied to diversified base learners and combined with other classical ensemble methods, such as Bagging. We incorporate Quantum-Inspired Subspace into Random Forest and propose Quantum-Inspired Forest. The additional computational cost is very cheap, equivalent to the cost of PCA preprocessing and linear regression. Second, we propose quantum interpretations for several machine learning concepts, and successfully establish a theoretical bridge between quantum interpretations and ensemble learning.

In future research, we consider introducing quantum entanglement to generate feature subsets. In quantum entanglement, the probability of selecting feature $|x_{1}\rangle$ is not independent of selecting feature $|x_{2}\rangle$. In the language of physics, at present, we have only used the mechanism of mixed states to design new methods. The concept of mixed states comes from quantum mechanics, but the density matrices of mixed states are always diagonal. It means the probabilistic distribution we use in present work is still classical probabilistic, although quantum physics helps us get it. We think it's accessible to apply superposition states and non-diagonal density matrices for constructing a non-classical probability distribution. In this approach, we can describe ensemble method based on the complex probability amplitude rather than the real probability. The relationship between probability amplitudes and the probabilities is same as the relationship between electromagnetic fields and light intensities. Amplitudes (wave) may intervene, while intensities may not. Overall, the application of non-diagonal density matrices can further make it possible to select features in a non-classical probabilistic distribution.

\bibliography{qiforest}

\end{document}